\documentclass[sn-mathphys-ay]{sn-jnl}

\usepackage{graphicx}%
\usepackage{multirow}%
\usepackage{amsmath,amssymb,amsfonts}%
\usepackage{amsthm}%
\usepackage{mathrsfs}%
\usepackage[title]{appendix}%
\usepackage{xcolor}%
\usepackage{textcomp}%
\usepackage{manyfoot}%
\usepackage{booktabs}%
\usepackage{algorithm}%
\usepackage{algorithmicx}%
\usepackage{algpseudocode}%
\usepackage{listings}%
\usepackage{lmodern}%
\usepackage{anyfontsize}
\usepackage[utf8]{inputenc}
\usepackage[english]{babel}

\DeclareMathOperator*{\argmax}{argmax}

\theoremstyle{plain}%
\theoremstyle{plain}%

\theoremstyle{plain}%
\newtheorem{definition}{Definition}%

\begin{document}

\title[Article Title]{Exploration in Knowledge Transfer Utilizing Reinforcement Learning}


\author*[1,2]{\fnm{Adam} \sur{Jedlička}}\email{jedlicka@utia.cas.cz}

\author[1]{\fnm{Tatiana Valentine} \sur{Guy}}\email{guy@ieee.org}


\affil[1]{\orgdiv{Institute of Information Theory and Automation}, \orgname{Czech Academy of Sciences}} 

\affil[2]{\orgdiv{Faculty of Applied Sciences}, \orgname{University of West Bohemia}}



\abstract{This article focuses on the problem of exploration within the task of knowledge transfer. Knowledge transfer refers to the useful application of the knowledge gained while learning the source task in the target task. The intended benefit of knowledge transfer is to speed up the learning process of the target task.
The article aims to compare several exploration methods used within a deep transfer learning algorithm, particularly Deep Target Transfer Q-learning. The methods used are $\epsilon$-greedy, Boltzmann, and Upper Confidence Bound exploration. The aforementioned transfer learning algorithms and exploration methods were tested on the virtual drone problem.
The results have shown that the Upper Confidence Bound algorithm performs the best out of these options. Its sustainability to other applications is to be checked.}




\maketitle
\newpage
\section{Introduction}\label{sec1}
 
The goal of this paper is to introduce an efficient method to transfer knowledge from a decision-making task with a known solution (source task) to another decision-making task with an unknown solution (target task). The decision-making tasks are modeled using the Markov decision process and solved using reinforcement learning (RL). 
Some of the problems one may encounter are a high dimension of the decision-making task, the mechanism of utilizing old knowledge in new tasks and the proper choice of the exploration method. The first problem is solved by RL modification called deep reinforcement learning. The remaining two problems form an objective of this paper.
\vspace{0.25cm}
\\* RL is one of the widely used methods for solving complex tasks, with the advantage of not requiring prior knowledge of the environment model. Some examples of RL are Q-learning actor-critic methods and the Proximal Policy Optimization algorithm. 
\\*However, these are suitable only for learning tasks with small state-action spaces. For solving tasks with large state-action spaces the aforementioned modification of RL into deep reinforcement learning (deep RL) is necessary. This modification uses neural networks to model the state-action spaces and saves the knowledge in the parameters of this neural network. 
\vspace{0.25cm}
\\*While RL methods are capable of converging to a solution of various tasks and solving them better than a human would, they bring some new challenges that need to be faced.
\vspace{0.25cm}
\\* One such challenge is finding a way to apply knowledge gained from learning one task (source task) to a similar task (target task), that differs in some way. The correct application of knowledge from the source task should achieve more efficient learning of the target task.
This problem is called knowledge transfer and the learning algorithm utilizing a correct knowledge transfer is called Transfer Learning (TL). 
An example of  TL can be seen in \cite{towards}, where an atari game called "boxing" is learned as a source task using deep Q-learning with experience replay. The knowledge from this game is then utilized in the target task represented by another atari game called "breakout".
\vspace{0.25cm}
\\*Another article discussing knowledge transfer is \cite{NetDuel}, where TL was applied to the problem of a virtual drone navigating through an environment based on images from a camera attached to it. 
However, \cite{towards} uses the parameters gained from learning from the source task as the initial parameters for learning the target task. If the considered tasks are not similar enough, then the learning process could be impaired. Likewise, \cite{NetDuel} states that its proposed method is sensitive to task similarity. 
 The problem of task similarity is addressed in \cite{repaint}, which uses the actor-critic method called Repaint. The Repaint method is based on proximal policy optimization and the article applies it to the problem of cars racing on a track. Repaint has positive results compared to baseline (non-TL) deep learning on similar tracks but outperforms baseline deep learning on widely different tracks.
\vspace{0.25cm}
\\*A TL approach used as the basis for this paper is outlined in \cite{Wang}. This approach modifies Q-learning with an additional decision rule between instances when the knowledge of the source task contained in its $Q$-function is useful for the target task solution and when it is not. This results in an algorithm named in \cite{Wang} as Target Transfer Q-learning (TTQL). Although this algorithm requires the tasks to have some level of similarity, the key advantage of this method is that the convergence is guaranteed \cite{Wang}. 
However \cite{Wang}, provides a TTQL algorithm suitable only for small state-action spaces. This paper introduces the generalization of this algorithm into a method that has deep RL as its cornerstone and can tackle problems with large state-action spaces.
\vspace{0.25cm}
\\*An additional challenge of RL addressed in this paper is the exploration problem. The exploration method refers to the strategy by which the state-action space is explored during the learning process. The efficiency of convergence of exploration methods is evaluated based on the RL algorithm and the solution. Some exploration methods face the danger of converging to non-optimal solutions while other exploration methods take too long to converge. The goal is to choose the exploration method with reasonable convergence time to a near-optimal solution. The exploration methods for RL are extensively discussed in \cite{amin} and more specifically on deep RL in \cite{ladosz}. The latter article also provides experimental results for various exploration strategies applied to MuJOCo and Atari games, which often serve as benchmark tasks. 
\vspace{0.25cm}
\\*Exploration is an important part of knowledge transfer, but the effects it has are not sufficiently tested. This article introduces a modified version of TTQL \cite{Wang} and compares the performance of several exploration methods. It is tested on a virtual drone task, where a drone navigates through a virtual room. This task is identical to the task in \cite{NetDuel} and thus code from this work was modified to meet the aims of this article.
\vspace{0.25cm}
\\*The paper layout is as follows. Section 2 defines the used notation, recalls the Markov Decision Process (MDP) and then sketches the basic theory behind RL, knowledge transfer and exploration. Section 3 applies the above-mentioned theory to a virtual drone. It is used for testing exploration in knowledge transfer
Section 4 presents the test results. Section 5 summarizes the results and outlines open problems.

\newpage
\section{Preliminaries}
\subsection{Notations and definitions}
The basic notations and definitions used throughout are in Table 1.
\begin{table*}[ht!]
\centering
\begin{tabular}{|l|l|l|}
\hline

\textbf{Name} & \textbf{Notation} & \textbf{Meaning}     

\\ \hline
Set of actions                                          & $\mathcal{A}$        &  a countable set of actions\\ \hline
Set of states                                          & $\mathcal{S}$        &  a countable set of observable states. 
\\ \hline
Time                                           & $t$        & \begin{tabular}[c]{@{}l@{}} discrete time labelled by integers  \end{tabular}  
\\ \hline
Action                                             & $a_t\in \mathcal{A}$        & \begin{tabular}[c]{@{}l@{}}an action at time $t$ \end{tabular}                                                                                                                                   \\ \hline
State                                              & $s_t\in \mathcal{S}$        & \begin{tabular}[c]{@{}l@{}}a state at time  $t$ \end{tabular}                                                                                                                                     \\ \hline
Policy rule                                              & $\pi:\mathcal{S} \rightarrow \mathcal{A}$       & \begin{tabular}[c]{@{}l@{}} $\pi(a_t\vert s_t)$ is a function assigning action $a_t$ \\ to state $s_t$. The policy is a time sequence \\of rules. \end{tabular}                                                                                                                                                \\ \hline
Discount factor                                     & $\gamma \in (0,1]$    & \begin{tabular}[c]{@{}l@{}} the weight reflecting the importance of future  \\ rewards\end{tabular}                                                               \\ \hline
\begin{tabular}[c]{@{}l@{}} Transition  \\ function
\end{tabular}      & $p:\mathcal{S} \times \mathcal{S} \times \mathcal{A} \rightarrow [0,1]$   & \begin{tabular}[c]{@{}l@{}}$p(s_{t+1}\vert s_t,a_t )$ is a probability of state $s_{t+1}$ \\ when applying action $a_t$ in state $s_t$, \\$\sum_{s_{t+1}\in\mathcal{S}} p(s_{t+1}\vert s_t,a_t) = 1$, $ \forall s_t\in\mathcal{S}, \forall a_t\in\mathcal{A}$  \end{tabular}       \\ \hline                                                                                                                                
\begin{tabular}[c]{@{}l@{}} Reward\\function\end{tabular}                                      & $r:\mathcal{S} \times \mathcal{S}\times \mathcal{A} \rightarrow R$        & \begin{tabular}[c]{@{}l@{}}Reward $  r(s_{t+1},s_t,a_t)$ is received when tran-\\sitioning from state  $s_t$ to state $s_{t+1}$ while \\ applying action $a_t$. Here, we distinguish \\ $r(s_{t+1},s_t,a_t)$ and $r_t$. The first one denotes\\ reward function  and the last denotes value \\ of the reward gained at time $t$.\end{tabular} \\ \hline
$Q$ - function                             & $Q^{\pi}:\mathcal{S}\times \mathcal{A}\rightarrow R$   & \begin{tabular}[c]{@{}l@{}}Function $Q^{\pi}(s_t,a_t)$ determines the value of \\state-action pair when policy $\pi$ is executed.\end{tabular}                                                                       \\ \hline
\begin{tabular}[c]{@{}l@{}} Discrete  \\ mean value
\end{tabular}                            & $ E[X] = \sum_{i \in I} p_i x_i $  & \begin{tabular}[c]{@{}l@{}}Mean value of discrete random variable $X$,\\ where $p_i$ is probability of $X = x_i$, $i\in I$ and \\ $x_i$ are values of $X$. \end{tabular}                                                                      \\ \hline
\end{tabular}
\caption{Notations and definitions} 
\end{table*}
\subsection{Markov Decision Process and Reinforcement Learning}
Markov decision process (MDP) is a mathematical formulation of a general decision-making process that suits to model tasks mentioned in the introduction. In such a process we define an agent as an entity that interacts with an environment, by observing the environment state and influencing it by action to reach the predefined goal. The degree to which the agent reaches his goal is measured by the reward the agent obtains after taking an action that changes the state of the environment. The new state depends on the particular action chosen and on the transition function. The value of the reward the agent obtains is determined by the reward function. 
\vspace{0.25cm}
\\*The basic scheme of this interaction is shown in Fig. 1. 
At time $t$ the agent observes environment state $s_t$ and chooses action $a_t$. The action and environment dynamics change environment from $s_{t}$ to $s_{t+1}$ and the agent receives reward $r_{t+1}$, quantifying the value of transition $(s_t,a_t) \rightarrow s_{t+1}$ for the agent.
\begin{figure}[ht!]
\centering
    \includegraphics[scale = 0.6]{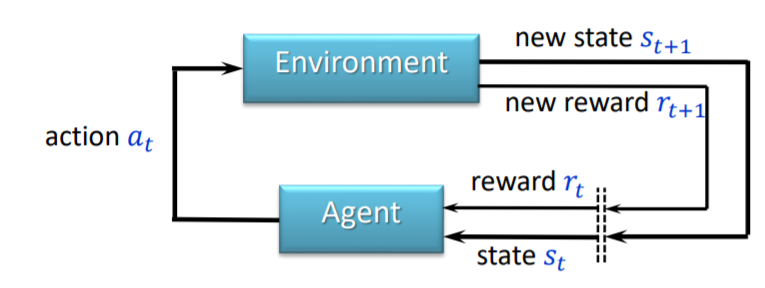}
    \caption{Interaction of the agent with the environment in MDP}
\end{figure}
vspace{0.25cm}
\\*What makes the decision process a Markov one is the Markov property. Process has Markov property when condition $p(s_{t+1}\vert a_{t},s_{t},, a_{t-1}, s_{t-1}...) = p(s_{t+1}\vert a_t,s_t)$ is satisfied. We assume that the environment follows Markov property. Then MDP is defined as a tuple $(\mathcal{S},\mathcal{A},p,\gamma,r)$, for notation see Table 1. Note that policies with Markov rules include the optimal policy.
\vspace{0.25cm}
\\*The agent has a goal to select policy $\pi:S \rightarrow A$ that maximizes its expected reward (mean value of reward gained by the time agent selects the last action). One possible way to find an optimal policy is with the help of $Q$-function. $Q$-function evaluates state-action $(s_t,a_t)$ pair for given policy $\pi$. It is formalized in Definition 1.\\
\vspace{0.25cm}
\begin{definition}
[$Q$-function] $Q$-function $Q^{\pi}(s,a)$ is defined as \cite{Wang} $Q^{\pi}(s,a) = E^{\pi}[\sum_{t=0}^{\infty} \gamma^t r(s_{t+1},s_t,a_t)\vert s_0 = s, a_0 = a]$, where $r(\cdot,\cdot,\cdot)$ is a reward function and $\gamma \in (0,1)$ is a discount factor. $E^{\pi}()$ refers to a conditional mean value of the argument over actions and states with fixed policy $\pi$.\\
\end{definition}
\vspace{0.25cm}
\noindent The $Q$-function function under the optimal policy (denoted with subscript $"opt"$)  obeys Bellman equation (1), which maximizes the overall accumulated reward
\begin{equation}
    Q^{opt}(s_t,a_t) =  \sum_{s_{t+1}\in \mathcal{S}}p(s_{t+1}\vert s_t,a_t)\left[ r(s_{t+1},s_t,a_t)  + \gamma \max_{a \in \mathcal{A}}Q^{opt}(s_{t+1},a)\right].
\end{equation}
\\*The rule $\pi^{opt}$ of the optimal policy follows the equation. See Table 1.
\begin{equation}
    \pi^{opt}(a_t\vert s_t) = a_t \in \argmax_{a\in\mathcal{A}} Q^{opt}(s_t,a). 
\end{equation}
When the transition function or reward function is not known to the agent, one of the possible approaches is to use RL. In RL, the agent learns the reward or transition functions based on the reward obtained by measuring the outcomes of actions. At the start of the learning process, when the agent is not familiar with the environment, he mostly takes random actions to explore new state-action pairs (exploration). When he gathers more knowledge he then exploits this knowledge to be closer to his goal (exploitation) and explores more sparsely.
\vspace{0.25cm}
\\*One such, widely used, RL algorithm is called Q-learning. This algorithm works with $Q$-function, Definition 1, and mimics the Bellman equation (1). For better understanding, we outline the first episode of Q-learning.
Before the episode starts the discrete time $t$, which serves as the algorithm's iterative index is set to 0. At the start of the first episode, the agent is placed into a certain starting state and given no knowledge about the reward or transition functions, i.e., the initial $Q$-function is full of zeros. Next, the agent takes an action. As mentioned earlier, the choice of this action is dependent on the exploration strategy. The state of the environment changes according to the chosen action and the transition function. Furthermore, the agent gets a reward by the environment based on the reward function and updates his knowledge, i.e., the $Q$-function. This iteratively goes on, until the agent reaches a predefined terminal state and the episode ends. 
Once this happens, the agent is put into the starting state, but now armed with knowledge from the previous episode in the form of a non-zero $Q$-function. The learning process stops by meeting a certain stopping rule. This rule is usually either reaching a certain maximum number of episodes, reaching the maximum number of time iterations or consistently reaching the goal in sufficiently short episodes,
The Q-learning algorithm updates the estimate of $Q$-function as the weighted average of the current value of $Q$-function estimate, $Q^{old}$, and the value of the maximum expected future reward \cite{DP}. The Q-function update equation reads
\begin{equation}
Q^{new}(s_t,a_t) = (1-\alpha) Q^{old}(s_t,a_t)  + \alpha\left[r_t + \gamma \max_{a \in \mathcal{A}}Q^{old}(s_{t+1},a) \right],
\end{equation}
\\*where $\alpha \in (0,1)$ is the optional learning rate. The higher its value, the higher the weight of newly gained information is. Generally, $\alpha$ can change with time but here we suppose it is fixed. 
The new knowledge about the task obtained by the agent is inserted into his current knowledge, represented by the  $Q^{old}$, with the term in the square brackets behind $\alpha$. Let us further refer to this term as an update term.
\subsection{Deep Q-learning}
In some cases, when the state space is too large, the basic Q-learning algorithm (3) is not viable due to computational complexity. In such cases deep Q-learning algorithm \cite{mnih} can be used. This method combines Q-learning with neural networks, which are used to estimate the value of the Q-function based on state and action input (Q-function network). The crucial advantage is that the knowledge about the Q-function (and solved task) is saved in the network parameters. Thus, unlike in regular Q-learning, it is not necessary to store the values of the Q-function for every state-action pair separately. The last property is the reason why this approach is, in practice, more suitable for large state-action spaces. Another common difference comes with the learning rate parameter $\alpha$ being removed (set to 1) from the update equation (3) and replaced by a similar parameter in neural network training (ADAM optimizer \cite{repaint}).
\vspace{0.25cm}
\\*Additionally, deep Q-learning also introduces new parameters, primarily experience replay, training interval, and target network \cite{mnih}. The parameters and their values are used in Experimental setup and results section in Table 2.
\vspace{0.25cm}
\\*Experience replay stores values $(s_{t+1},a_t,s_t,r_t)$ into a replay memory structure. During the training the random batch of a certain size is chosen from this replay memory structure to train on. This helps to break possible correlation that often occurs between consecutive transitions. 
\vspace{0.25cm}
\\* Training interval refers to a number of iterations during which the Q-function network stays the same and after which the training occurs using the batch chosen from replay memory.
\vspace{0.25cm}
\\* The target network introduces an additional neural network, that is not updated every training interval and it is a copy of the Q-function network. This network is used to estimate the Q-function value of the next state in the update equation (3). This provides a more stable target for the Q-function updates. After a certain amount of iterations (target network update period) the Q-function network and target network parameters are averaged. The resulting network serves as a starting point for the new target and Q-function network for the next target networks update period.

\subsection{Knowledge transfer and target transfer Q-learning}
Knowledge transfer refers to using knowledge gained by learning one task defined by an MDP (source task), to another, different task defined by a different MDP (target task). The aim of this transfer is to significantly speed up the convergence to the solution of the target task. This can be formulated formally in the following way \cite{Barreto}.
\vspace{0.25cm}
\begin{definition}
Let us have two sets of tasks $\mathcal{C}_1$, and $\mathcal{C}_2$ such, that $\mathcal{C}_1 \subset \mathcal{C}_2$. Now we consider any task $T$. We say that there is a beneficial knowledge transfer if, after training the agent for task $T$ on $\mathcal{C}_2$, it performs equally well or better than if only trained on $\mathcal{C}_1$.
\end{definition}
\vspace{0.25cm}
\noindent A learning that utilizes knowledge transfer is called transfer learning (TL).
\vspace{0.25cm}
\\* One of the algorithms, that modifies regular Q-learning into TL learning is called Target transfer Q-learning (TTQL) \cite{Wang}. As the name suggests, this algorithm is similar to the simple Q-learning mentioned earlier. The key difference is, that TTQL introduces a decision rule on when to use the knowledge from the source task and when not. The algorithm is described in detail in \cite{Wang}, but the mechanism of how the knowledge is transferred can be explained by the following line of thought.
\vspace{0.25cm}
\\*Let us imagine a situation, where we have a task and its solution, i.e. the optimal $Q$-function, available. Let us further suppose that we want to design an intuitive way of implementing the knowledge of the solution of the task into the $Q$-learning algorithm. If we recall update equation (3), it was mentioned that new knowledge is inserted into $Q$-function updated with the help of an update term. Therefore, the most intuitive way would be to substitute $Q^{old}$ with the optimal $Q$-function into the update term. 
This situation can be also interpreted as if we had a source task and a target task, which are both exactly the same. This means that such a situation could be interpreted as a very special case of knowledge transfer.
\vspace{0.25cm}
\\*Of course, the problem is that the knowledge transfer between two exactly the same tasks is not very useful. Let us now assume that the source task and the target task are different, but may share some similarities. We can assume that the more similar such tasks are the more similar in values their optimal $Q$-functions will be. Let us denote the optimal $Q$-function of the target task as $Q^{opt}_{target}$ and the optimal $Q$-function of the source task as $Q^{opt}_{source}$. This means, that if there is more similarity between the $Q^{opt}_{target}$ and $Q^{opt}_{source}$ than between the $Q^{opt}_{target}$  and the current estimate of the $Q^{opt}_{target}$ it would be desirable to substitute $Q^{opt}_{source}$ into update term in update equation (3).
This raises a question of how can the agent decide when the similarity is sufficient. This is done by decision rule derived in detail in \cite{Wang}. 
\vspace{0.25cm}
\\*The summary of the derivation is as follows. The similarity between the two tasks can be defined as the maximal distance of their optimal Q-functions defined as (4).
\begin{equation}
 \Delta(Q^{opt}_{source}, Q^{opt}_{target}) = \max_{\substack{a_t \in \mathcal{A} \\ s_t \in \mathcal{S}}} \vert Q^{opt}_{source}(a_t, s_t) - Q^{opt}_{target}(a_t, s_t)\vert
\end{equation}
Following the same principle the similarity between $Q^{opt}_{target}$ and its current non-updated estimate $Q^{old}$ can be written as (5).
\begin{equation}
 \Delta(Q^{old}, Q^{opt}_{target}) = \max_{\substack{a_t \in \mathcal{A} \\ s_t \in \mathcal{S}}} \vert Q^{old}(a_t, s_t) - Q^{opt}_{target}(a_t, s_t)\vert
\end{equation}
The ideal decision rule proposed in \cite{Wang} would compare the results of equations (4) and (5). In case $\Delta(Q^{opt}_{source}, Q^{opt}_{target})$ is smaller than $\Delta(Q^{old}, Q^{opt}_{target})$, $Q^{opt}_{source}$ is used in the update term and knowledge transfer takes place. In the opposite case, $Q^{old}$ is used and regular Q-learning iteration takes place. However, since $Q^{opt}_{target}$ is unknown this rule cannot be used. Therefore \cite{Wang} further proposes to use Bellman error defined as (6)
\begin{equation}
 MNBE(Q(s_t,a_t)) = \max_{\substack{a_t \in \mathcal{A} \\ s_t \in \mathcal{S}}} \vert Q(s_t,a_t) - (r_t + \gamma E_{s_{t+1}} \max_{a_{t+1}\in \mathcal{A}}Q(a_{t+1},s_{t+1}))\vert,
\end{equation}
where symbol $E_{s_{t+1}}$ represents conditional mean value over the new state $s_{t+1}$. Bellman error is then used in the decision rule as a replacement for both  $\Delta(Q^{opt}_{source}, Q^{opt}_{target})$ and $\Delta(Q^{old}, Q^{opt}_{target})$. The correctness of this is derived in \cite{Wang}, where it is proved following relationship 
\begin{equation}
\Delta(Q^{old}, Q^{opt}_{target}) \leq \frac{MNBE(Q^{old})}{1-\gamma}.
\end{equation}
Similar relationship (8) for $\Delta(Q^{opt}_{source}, Q^{opt}_{target})$ has not been proven in \cite{Wang}, but a trivial modification of the mentioned proof yields the proof of 
\begin{equation}
\Delta(Q^{opt}_{source}, Q^{opt}_{target}) \leq \frac{MNBE(Q^{opt}_{source})}{1-\gamma}.
\end{equation}
It is important to note, that $Q^{opt}_{source}$ is in practice not known perfectly, as the result of learning the source task using a regular Q-learning is only an estimate of $Q^{opt}_{source}$. In practice, this estimate denoted also $Q^{opt}_{source}$ is used. For completion, the final decision rule is as follows. 
In case $MNBE(Q_{source})$ is smaller than $MNBE(Q_{old})$, $Q_{source}$ is used in the update term and knowledge transfer takes place. In the opposite case, $Q^{old}$ is used and regular Q-learning iteration takes place.
An unexpected property of this result is that the decision rule uses $MNBE(Q_{source})$, which does not explicitly depend on the Q-function of the target task in any way. This is explained by the reward term in equation (6), which connects $MNBE(Q_{source})$ to the target task by taking reward values measured during the current learning process of the target task into account.
\vspace{0.25cm}
\\* As in the case of Q-learning, the TTQL needs modification when dealing with large state spaces. This modification will be called deep TTQL throughout the article. It is again done with the help of a neural network that estimates the $Q$ function based on collected data.
In this modification, the knowledge about the source task is contained in the parameters of the neural network trained while learning the source task (Source Q-function network). Similarly, the gathered target task knowledge is contained in the parameters of the neural network dedicated to the target task (Main Q-function network). Additionally, the Target network can be added to the Main Q-function network as outlined in deep Q-learning, which can help to stabilize learning. Target network and target task are not associated in any way, but unfortunately naming conventions make them appear as related.
The TTQL decision rule then uses these estimates of $Q$-function in (6) instead of the exact values used in basic TTQL.
\subsection{Exploration}
This section focuses on balancing exploration and exploitation mentioned in the introduction. Intuitively, a good strategy would prefer to explore new state-action pairs more often at the start of the learning process to gather more knowledge and to explore less later on, while choosing (exploiting) the already discovered high-performing state-action pairs. In other words, the agent will gather up some knowledge and then utilize this knowledge to be close to the optimal policy. The exploration strategies can be implemented using the various exploration methods that determine a portion of exploration and exploitation based on the current knowledge at time $t$ in the learning process. 
\vspace{0.25cm}
\\*The exploration methods tested by us use $\theta(a_t\vert s_t)$, which is the probability of taking action $a_t$ given we are at state $s_t$.  Note, that such an exploration method is not a replacement for the transition function as it might seem. The transition function is an inherent property of  MDP while exploration methods are used to efficiently balance exploration and exploitation.
\vspace{0.25cm}
\\*The first and simplest method is $\epsilon$-greedy method (9). In it, $A$ denotes the total number of available actions at state $s_t$. Parameter $\epsilon_t>0$ lowers either linearly or exponentially with time from a predefined starting value close to 1 to a predefined minimum value close to 0. The exploration probability is
\begin{equation}
        \theta(a_t\vert s_t) =  \begin{cases} 1 - \epsilon_t & \text{if}\; a_t =  \argmax_{a\in\mathcal{A}} Q(s_t,a) \\\frac{\epsilon_t}{A - 1}
                                                          & \text{otherwise}.     %
        \end{cases}
\end{equation}
\\* The upper line in (9) corresponds to exploitation and the lower line to exploration. We can see that as parameter $\epsilon_t$ decreases the probability of taking an action that exploits the current knowledge increases.
\vspace{0.25cm} 
\\*The other method discussed here is the Boltzmann method. This exploration method constructs Boltzmann probability distribution (6) based on the value of the $Q$-function and parameter $\lambda_t>0$.
\begin{equation}
     \theta(a_t\vert s_t) = \frac{e^(\frac{Q(s_t,a_t)}{\lambda_t})}{\sum_{a_t \in \mathcal{A}}e^(\frac{Q(s_t,a_t)}{\lambda_t})}.
\end{equation}
 We can see, that for large $\lambda_t$ this distribution is nearly uniform. This motivates the choice of high $\lambda_t$ at the start, so all actions have similar probability and exploration is maximized. Over time $\lambda_t$ is lowered to close to 0 and the probability of taking the most beneficial action, i.e., exploiting the knowledge, increases. 
 \vspace{0.25cm}
\section{Tested knowledge transfer with exploration}
The above-mentioned methods were implemented on a virtual drone problem. This handles a virtual drone (modeled in Unreal Engine 4 software) navigating through a room and avoiding obstacles. This section shows how the MDP and knowledge transfer terminology outlined earlier is implemented in this task. The source task is a drone navigating according to a certain virtual map based on its camera images and the target task is the same drone navigating a different virtual map. The implementation is modeled after \cite{NetDuel}, which offers the framework for classical deep Q-learning. An addition to the implementation in \cite{NetDuel} is the ability of the drone to use different exploration methods and transfer knowledge from previously learned tasks. 
\vspace{0.25cm}
\\* The states are represented by images from an RGB camera (camera with colored images) attached to the drone
These images are then resized to reduce the complexity. The initial state is fixed for each map and the terminal state is represented as any state where the drone hits an obstacle. Upon hitting an obstacle drone resets back to its initial state and starts a new episode of training.
\begin{figure}[ht!]
\centering
    \includegraphics[scale = 1]{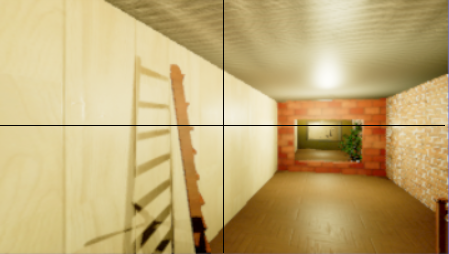}
    \caption{Image from RGB camera segmented by a grid into cells corresponding to possible actions}
\end{figure}
\vspace{0.25cm}
\\* Actions are defined in the following way. We split the image into a preset number of cells (see Fig. 2) and defined an action as moving straight toward the center of one of these cells with a constant speed for a time equal to the period between iterations. To get closer to a real situation, where drone movement is influenced by noise caused by various conditions. We model that noise by adding an artificial random effect into the action of the drone. This means that we introduce a small error to the speed and angle of movement of the drone when action $a_t$ is chosen.
 \begin{figure}[ht!]
\centering
    \includegraphics[scale = 1.03]{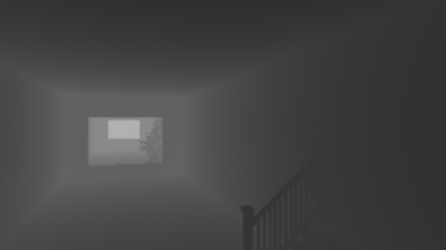}
    \caption{Example of an image from depth of field camera}
\end{figure}
\vspace{0.25cm}
\\* The virtual drone is also equipped with a depth of field camera (DoF), which records a greyscale image that shows closer objects darker and further objects brighter (Fig.3). The DoF camera is useful for modeling the reward, as the reward should be higher in places where the drone cannot see any nearby obstacle and lower in places where an obstacle is near.
The reward is calculated DoF image given at each state given by formulas taken from the source code of \cite{NetDuel} or can be seen in \cite{DP}.
\vspace{0.25cm}
\\* Neural network is modeled as in \cite{NetDuel}, as it was found to be very effective for the virtual drone problem. In the case of the source task, the parameters of the Q-function network are initialized with preset initial values used as a starting point. In the case of the target task, we deal with 2 versions of the neural network earlier denoted as Main Q-function network and Source Q-function network (See subsection 2.4). The first one is initialized to the same preset starting values as it was initialized at the start of training of the source task and serves as the main neural network of the task. The second one serves as a neural network used for knowledge transfer from the source task to the target task. The Source Q-function network has parameters initialized with the same values as the Q-function network of the source task after the training of the source task was finished, i.e., the output values of the training of the source task. The amount of predictors (neural network inputs) is dependent on the size of the input image. The input image is taken as a cutout of the image around the center of the camera of a predefined width and height. This cutout can be potentially scaled up to the full image. In our case, we used input images 103 pixels tall and 103 pixels wide.
\vspace{0.25cm}
\\*The scheme of the used neural network is shown in Fig. A1 in Appendix A and the scheme of the overall deep TTQL algorithm applied to the virtual drone problem is shown in Fig. B1 in Appendix B.

\newpage
\section{Experimental setup and results}\label{sec2}
Multiple experiments were conducted to test the robustness of deep TTQL. As a source task for a drone, the navigating (i.e. flying through the map avoiding obstacles) through the same map (map name - pyramid) was chosen for every experiment. This source task was learned using deep Q-learning and knowledge about this task was saved in the weights of the trained neural network. Target tasks for the drone were chosen to be navigating through the different maps. The experiments differ by different maps chosen and by changes in learning parameters. The baseline experiment used parameters summarised in Table 2. The learning parameters are explained in more detail in the Reinforcement learning and deep learning subsections and exploration parameters in subsection 2.5. 
\vspace{0.25cm}
\\*To summarize the benefits of TTQL we will compare TTQL cases (in which we are using various exploration methods) with cases where we use no TL at all to learn the target task (i.e. we use simple deep Q-learning). 
\vspace{0.25cm}
\\*Additionally, we compare TTQL to cases where we use the Source Q-function network parameters as initial parameters of the Q-function network of the target task. In other words, we use simple deep Q-learning to learn the target task, but we preset initial Q-function network parameters to values that should be closer to the solution. This is done to show that the source task and the target task are different enough to have a reason for starting training on the target task with knowledge transfer as opposed to just continuing training the source task further.
\begin{table}[ht!]
\centering
\begin{tabular}{|l|l|}
\hline
\textbf{Parameter}             & \textbf{Value}                                                                                                                                                                                                                      \\ \hline

number of actions (cells image is divided into as described in Section 3)          & 25                                                                                                                                            \\ \hline
number of movements (iterations) the drone does before training of the network starts   & 5000                                                                                                                             \\ \hline
maximum number of iterations            & 150 000                                                                                                                                                                                                    \\ \hline

number of iterations after which $\epsilon$ in $\epsilon$ greedy method gets to its fixed lowest value  & 120 000                                                                              \\ \hline
discount factor (See Subsection 2.2)                & 0.99                                                                                                                                                                                                                    \\ \hline
dropout rate used in neural networks         & 0.1                                                                                                                                                                \\ \hline
learning rate (See Subsection 2.3)       & 0.000002                                                                                                                                                                                                            \\ \hline
training interval (See Subsection 2.3)        & 16                                                                                                                                                   \\ \hline
target network update period (See Subsection 2.3        & 15000                                                                                                                      \\ \hline
batch size (See Subsection 2.3)       & 32                                                                                                                                                                                                            \\ \hline
starting value of parameter $\lambda$ in Boltzmann exploration (See Subsection 2.5) & 1                                                                                                                                                                                      \\ \hline
final value of parameter $\lambda$ in Boltzmann exploration (See Subsection 2.5)       & 0.07    
            \\ \hline
starting value of $\epsilon$ parameter in $\epsilon$ greedy exploration (See Subsection 2.5)    & 1 \\ \hline
final value of $\epsilon$ parameter in $\epsilon$ greedy exploration (See Subsection 2.5)    & 0.05 \\ \hline
the loss function used for training neural network     &  MSE loss \\ \hline
\end{tabular}
\caption{Values of parameters in baseline Experiment 1 (Map name - Techno).}
\end{table}
\\*The following experiments were conducted.
\begin{itemize}
    \item Experiment 1, map name - techno, baseline experiment 
    \item Experiment 2, map name - complex, comparison of the number of knowledge transfer instances evolving with time for different exploration methods.
\end{itemize}
The results of these experiments can be seen in Fig. 4 and Fig. 5.
\vspace{0.25cm}
\\*If we recall Section 3, a single episode consists of the number of iterations in which the drone did not crash into an obstacle i.e., the episode starts when the drone starts in the initial position and ends when the drone crashes into an obstacle. 
\vspace{0.25cm}
\\*Since the results of the experiment measured reward accumulated over an episode, each graph in the chart ends at a different point. This is due to the fixed number of total time iterations for each experiment and episodes taking a variable number of time iterations to end. 
For example, at the start, the drone has very little knowledge collected so it may crash after several time iterations and end the episode quickly. On the other hand, once it has collected some knowledge an episode between start and crash will take more time iterations.
\vspace{0.25cm}
\\*Stopping the learning process after a fixed amount of iterations is done so we can measure the speed of learning and compare different exploration methods and the effectiveness of knowledge transfer. Note, that the task is formulated in such a way that cumulative reward does not converge to a certain value. This is because that drone with knowledge of the optimal solution would never hit an obstacle and thus it would keep accumulating more reward.
\vspace{0.25cm}
\\*Table 3 explains details behind labels used in graphs describing the results of the experiments \cite{DP}.

\begin{table}[ht!]
\centering
\begin{tabular}{|l|l|}
\hline
\textbf{Label}                      & \textbf{Explanation}                                                                                                                                                       \\ \hline
no transfer linear $\epsilon$ - greedy & \begin{tabular}[c]{@{}l@{}}deep Q-learning (with no transfer) with $\epsilon$-greedy exploration \\ algorithm and linear epsilon decrease\end{tabular} \\ \hline
no transfer exponential $\epsilon$ - greedy & \begin{tabular}[c]{@{}l@{}}deep Q-learning (with no transfer) with $\epsilon$-greedy exploration \\ algorithm and exponential epsilon decrease\end{tabular} \\ \hline
no transfer Boltzmann & \begin{tabular}[c]{@{}l@{}}deep Q-learning (with no transfer) with Boltzmann\\ exploration 
\end{tabular} \\ \hline
no TTQL linear $\epsilon$ - greedy & \begin{tabular}[c]{@{}l@{}}deep Q-learning with initial Q-network parameters set \\ as parameters of Q-network learned in the source task \\  with  $\epsilon$-greedy exploration with linear decrease 
\end{tabular} \\ \hline
no TTQL exponential $\epsilon$ - greedy & \begin{tabular}[c]{@{}l@{}}deep Q-learning with initial Q-network parameters set \\ as parameters of Q-network learned in the source task \\  with  $\epsilon$-greedy exploration with exponential decrease 
\end{tabular} \\ \hline
no TTQL Boltzmann & \begin{tabular}[c]{@{}l@{}}deep Q-learning with initial Q-network parameters set \\ as parameters of Q-network learned in the source task \\  with Boltzmann exploration 
\end{tabular} \\ \hline
TTQL linear $\epsilon$ - greedy         & \begin{tabular}[c]{@{}l@{}}Target Transfer Q-learning with $\epsilon$-greedy exploration \\ algorithm and linear epsilon decrease \end{tabular}              \\ \hline
TTQL exponential $\epsilon$ - greedy    & \begin{tabular}[c]{@{}l@{}}Target Transfer Q-learning with $\epsilon$-greedy exploration \\ algorithm and exponential epsilon decrease \end{tabular}         \\ \hline
TTQL Boltzmann                     & \begin{tabular}[c]{@{}l@{}}Target Transfer Q-learning with Boltzmann exploration \\algorithm \end{tabular}                                                                                                          \\ \hline

\end{tabular}
\caption{Exploration methods used in experiments}
\end{table}
\vspace{0.25cm}

\newpage
\begin{figure}[ht!]
\centering
    \includegraphics[scale = 0.67]{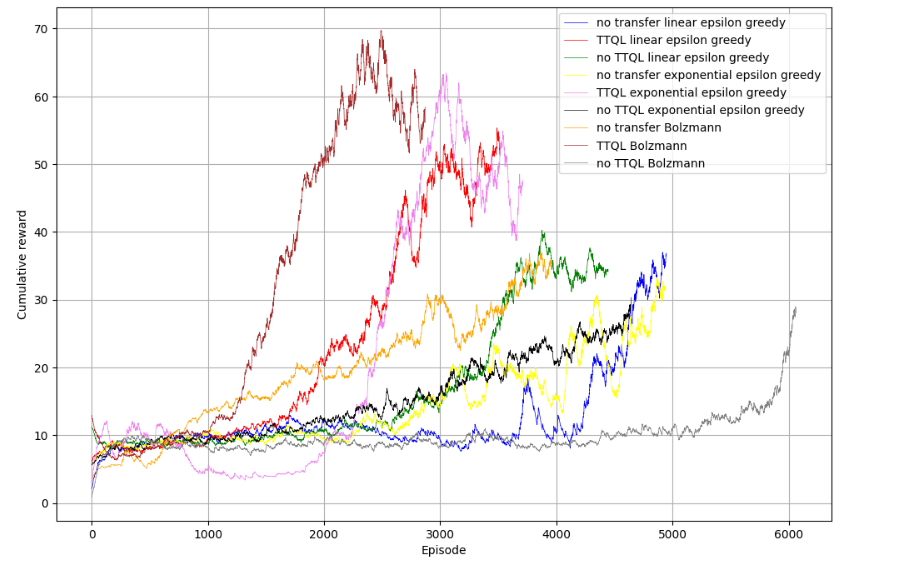}
    \caption{Results of Experiment 1 - Reward accumulated over episodes with different numbers of iterations}
\end{figure}
\begin{figure}[ht!]
\centering
    \includegraphics[scale = 0.67]{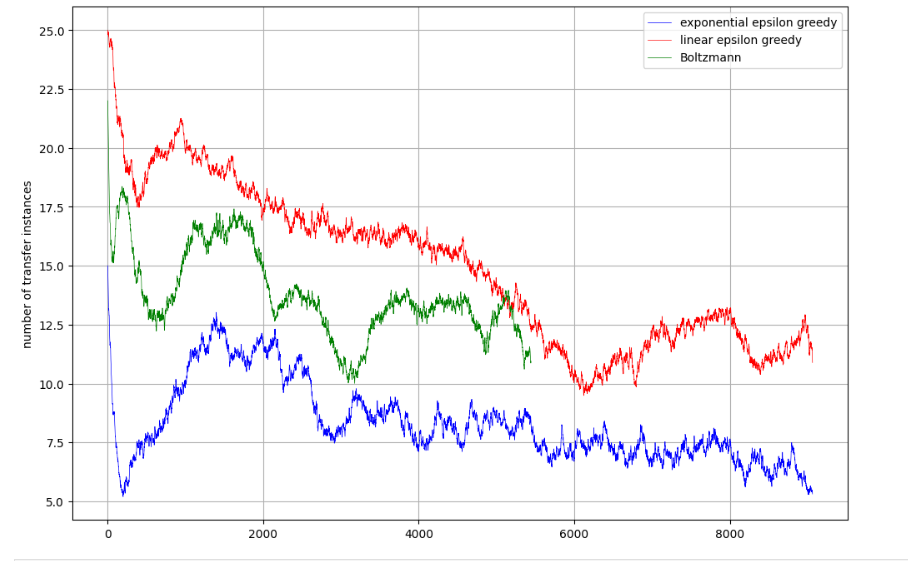}
    \caption{The evolution of transfer instances over time}
\end{figure}
\newpage
\section{Conclusion}
From Fig.4. we can see that the TTQL knowledge transfer method outperformed deep Q-learning with no transfer. This means that the knowledge transfer is beneficial and works as desired. 
\vspace{0.25cm}
\\*Furthermore, TTQL compared well to deep Q-learning with initial Q-network parameters set as parameters of Q-network learned in the source task (noTTQL). This shows that the source and target tasks are different enough to have meaningful knowledge transfer. In case the source and target task are extremely similar we would on the other hand expect the difference between these two methods to be minimal and this did not happen.
\vspace{0.25cm}
\\*As for the individual exploration methods, their performance relative to each other seems to vary based on whether TTQL, noTTQL, or no transfer at all are used. A general trend we can observe is that the Boltzmann method uses more exploitation and thus has a faster increase at the start as opposed to the $\epsilon$ - greedy methods. Another method called Upper confidence bound (UCB) was tested earlier in \cite{DP}. However, it was found that this method is not suitable for this task due to performing exploitation almost entirely and thus giving misleading results in previously mentioned work.
\vspace{0.25cm}
\\* In Fig. 5 we can see the evolution of transfer instances in the TTQL method. This graph shows us that the number of instances of knowledge transfer from the source task decreases as the drone gathers more knowledge, which is as expected.
\\* Future work will focus on reaching a better source task solution that is closer to the optimum for TL and 
testing a wider group of exploration methods on a wider group of tasks. For instance, navigating in the environment in a different way e.g. looking for a certain object or taking a trajectory of a certain shape.
\medskip



\nocite{*}
\bibliography{bibliography}


\begin{thebibliography}{9}
\ifx \bisbn   \undefined \def \bisbn  #1{ISBN #1}\fi
\ifx \binits  \undefined \def \binits#1{#1}\fi
\ifx \bauthor  \undefined \def \bauthor#1{#1}\fi
\ifx \batitle  \undefined \def \batitle#1{#1}\fi
\ifx \bjtitle  \undefined \def \bjtitle#1{#1}\fi
\ifx \bvolume  \undefined \def \bvolume#1{\textbf{#1}}\fi
\ifx \byear  \undefined \def \byear#1{#1}\fi
\ifx \bissue  \undefined \def \bissue#1{#1}\fi
\ifx \bfpage  \undefined \def \bfpage#1{#1}\fi
\ifx \blpage  \undefined \def \blpage #1{#1}\fi
\ifx \burl  \undefined \def \burl#1{\textsf{#1}}\fi
\ifx \doiurl  \undefined \def \doiurl#1{\url{https://doi.org/#1}}\fi
\ifx \betal  \undefined \def \betal{\textit{et al.}}\fi
\ifx \binstitute  \undefined \def \binstitute#1{#1}\fi
\ifx \binstitutionaled  \undefined \def \binstitutionaled#1{#1}\fi
\ifx \bctitle  \undefined \def \bctitle#1{#1}\fi
\ifx \beditor  \undefined \def \beditor#1{#1}\fi
\ifx \bpublisher  \undefined \def \bpublisher#1{#1}\fi
\ifx \bbtitle  \undefined \def \bbtitle#1{#1}\fi
\ifx \bedition  \undefined \def \bedition#1{#1}\fi
\ifx \bseriesno  \undefined \def \bseriesno#1{#1}\fi
\ifx \blocation  \undefined \def \blocation#1{#1}\fi
\ifx \bsertitle  \undefined \def \bsertitle#1{#1}\fi
\ifx \bsnm \undefined \def \bsnm#1{#1}\fi
\ifx \bsuffix \undefined \def \bsuffix#1{#1}\fi
\ifx \bparticle \undefined \def \bparticle#1{#1}\fi
\ifx \barticle \undefined \def \barticle#1{#1}\fi
\bibcommenthead
\ifx \bconfdate \undefined \def \bconfdate #1{#1}\fi
\ifx \botherref \undefined \def \botherref #1{#1}\fi
\ifx \url \undefined \def \url#1{\textsf{#1}}\fi
\ifx \bchapter \undefined \def \bchapter#1{#1}\fi
\ifx \bbook \undefined \def \bbook#1{#1}\fi
\ifx \bcomment \undefined \def \bcomment#1{#1}\fi
\ifx \oauthor \undefined \def \oauthor#1{#1}\fi
\ifx \citeauthoryear \undefined \def \citeauthoryear#1{#1}\fi
\ifx \endbibitem  \undefined \def \endbibitem {}\fi
\ifx \bconflocation  \undefined \def \bconflocation#1{#1}\fi
\ifx \arxivurl  \undefined \def \arxivurl#1{\textsf{#1}}\fi
\csname PreBibitemsHook\endcsname

\bibitem[\protect\citeauthoryear{Amin et~al.}{2021}]{amin}
\begin{botherref}
\oauthor{\bsnm{Amin}, \binits{S.}},
\oauthor{\bsnm{Gomrokchi}, \binits{M.}},
\oauthor{\bsnm{Satija}, \binits{H.}},
\oauthor{\bsnm{Hoof}, \binits{H.}},
\oauthor{\bsnm{Precup}, \binits{D.}}:
A survey of exploration methods in reinforcement learning
(2021)
\doiurl{10.48550/arXiv.2109.00157}
\end{botherref}
\endbibitem

\bibitem[\protect\citeauthoryear{Anwar and Raychowdhury}{2020}]{NetDuel}
\begin{barticle}
\bauthor{\bsnm{Anwar}, \binits{A.}},
\bauthor{\bsnm{Raychowdhury}, \binits{A.}}:
\batitle{Autonomous navigation via deep reinforcement learning for resource
  constraint edge nodes using transfer learning}.
\bjtitle{IEEE Access}
\bvolume{8},
\bfpage{26549}--\blpage{26560}
(\byear{2020})
\doiurl{10.1109/ACCESS.2020.2971172}
\end{barticle}
\endbibitem

\bibitem[\protect\citeauthoryear{Barreto et~al.}{2016}]{Barreto}
\begin{botherref}
\oauthor{\bsnm{Barreto}, \binits{A.}},
\oauthor{\bsnm{Munos}, \binits{R.}},
\oauthor{\bsnm{Schaul}, \binits{T.}},
\oauthor{\bsnm{Silver}, \binits{D.}}:
Successor Features for Transfer in Reinforcement Learning.
https://doi.org/10.48550/arXiv.1606.05312
(2016)
\end{botherref}
\endbibitem

\bibitem[\protect\citeauthoryear{Glatt et~al.}{2016}]{towards}
\begin{botherref}
\oauthor{\bsnm{Glatt}, \binits{R.}},
\oauthor{\bsnm{Silva}, \binits{F.}},
\oauthor{\bsnm{Costa}, \binits{A.}}:
Towards knowledge transfer in deep reinforcement learning
(2016)
\doiurl{10.1109/BRACIS.2016.027}
\end{botherref}
\endbibitem

\bibitem[\protect\citeauthoryear{Jedlička and Guy}{2023}]{DP}
\begin{botherref}
\oauthor{\bsnm{Jedlička}, \binits{A.}},
\oauthor{\bsnm{Guy}, \binits{T.}}:
Exploration in Knowledge Transfer.
https://dspace.cvut.cz/handle/10467/107197
(2023)
\end{botherref}
\endbibitem

\bibitem[\protect\citeauthoryear{Ladosz et~al.}{2022}]{ladosz}
\begin{botherref}
\oauthor{\bsnm{Ladosz}, \binits{P.}},
\oauthor{\bsnm{Weng}, \binits{L.}},
\oauthor{\bsnm{Kim}, \binits{M.}},
\oauthor{\bsnm{Oh}, \binits{H.}}:
Exploration in deep reinforcement learning: A survey
(2022)
\doiurl{10.48550/arXiv.2205.00824}
\end{botherref}
\endbibitem

\bibitem[\protect\citeauthoryear{Mnih et~al.}{2013}]{mnih}
\begin{botherref}
\oauthor{\bsnm{Mnih}, \binits{V.}},
\oauthor{\bsnm{Kavukcuoglu}, \binits{K.}},
\oauthor{\bsnm{Silver}, \binits{D.}},
\oauthor{\bsnm{Graves}, \binits{A.}},
\oauthor{\bsnm{Antonoglou}, \binits{D.} \bsuffix{Ioannis amd~Wierstra}},
\oauthor{\bsnm{Riedmiller}, \binits{M.}}:
Playing atari with deep reinforcement learning
(2013)
\doiurl{10.48550/arXiv.1312.5602}
\end{botherref}
\endbibitem

\bibitem[\protect\citeauthoryear{Tao et~al.}{2020}]{repaint}
\begin{botherref}
\oauthor{\bsnm{Tao}, \binits{Y.}},
\oauthor{\bsnm{Genc}, \binits{S.}},
\oauthor{\bsnm{Chung}, \binits{J.}},
\oauthor{\bsnm{Sun}, \binits{T.}},
\oauthor{\bsnm{Mallya}, \binits{S.}}:
REPAINT: Knowledge Transfer in Deep Reinforcement Learning.
International Conference on Machine Learning,
  https://doi.org/10.48550/arXiv.2011.11827
(2020)
\end{botherref}
\endbibitem

\bibitem[\protect\citeauthoryear{Wang et~al.}{2020}]{Wang}
\begin{botherref}
\oauthor{\bsnm{Wang}, \binits{Y.}},
\oauthor{\bsnm{Liu}, \binits{Y.}},
\oauthor{\bsnm{Chen}, \binits{W.}},
\oauthor{\bsnm{Ma}, \binits{Z.-M.}},
\oauthor{\bsnm{Liu}, \binits{T.-Y.}}:
Target transfer q-learning and its convergence analysis.
Neurocomputing
\textbf{392}
(2020)
\doiurl{10.48550/arXiv.1312.5602}
\end{botherref}
\endbibitem

\end{thebibliography}
\newpage
\begin{appendices}
\section{Neural network architecture}
 \begin{figure}[ht!]
\centering
    \includegraphics[height = 150mm,width = 140mm]{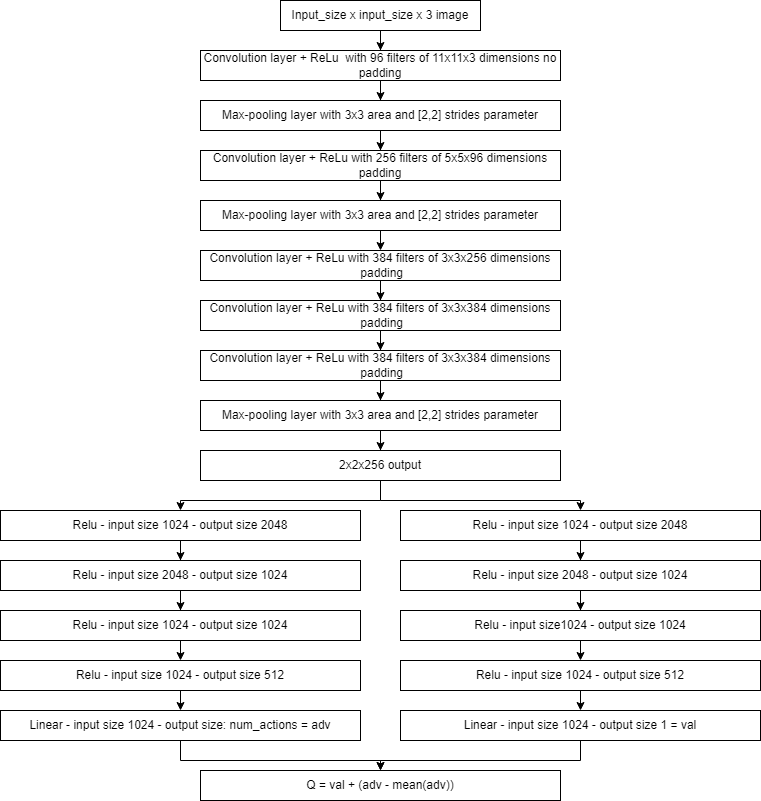}
    \caption{Network architecture}
\end{figure}
\newpage
\section{Overall algorithm structure}
\begin{figure}[ht!]
\centering
    \includegraphics[height = 170mm,width = 140mm]{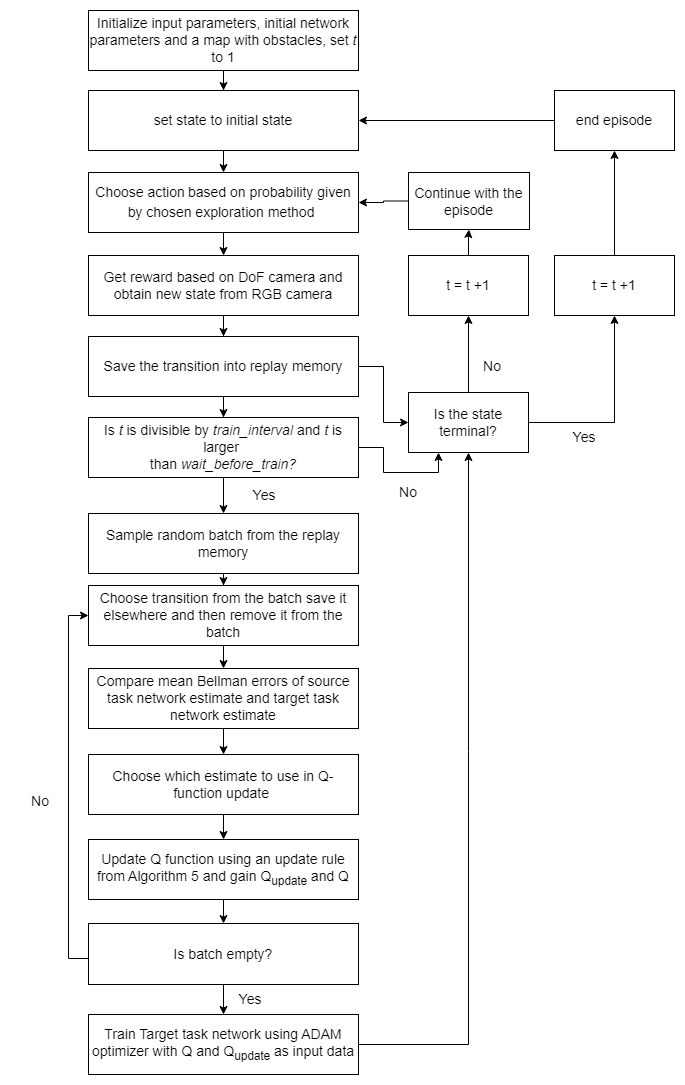}
    \caption{The overall scheme of the deep TTQL algorithm}
\end{figure}

\end{appendices}

\end{document}